\documentclass[letterpaper, 10pt, conference]{ieeeconf} 

\IEEEoverridecommandlockouts                              
\makeatletter
\let\NAT@parse\undefined
\makeatother
\usepackage[numbers]{natbib}
\usepackage{times, amsmath, amssymb}

\usepackage{amsthm}

\usepackage{verbatim, enumerate, color, url, graphicx, hyperref}
\usepackage{mathtools, wrapfig, epsfig, multirow, paralist}
\usepackage[caption=false]{subfig}
\usepackage{xcolor, comment, algorithm, algorithmic, lipsum}

\let\labelindent\relax
\usepackage{enumitem}

\newcommand{\x}{\ensuremath{\mathbf x}}
\newcommand{\y}{\ensuremath{\mathbf y}}

\newcommand{\h}{\ensuremath{\mathbf h}}

\newcommand{\W}{\ensuremath{\mathbf W}}

\newcommand{\tx}{\ensuremath{\tilde{\mathbf x}}}

\newcommand{\tf}{\ensuremath{\tilde{f}}}
\newcommand{\uda}{\ensuremath{\mathcal{U}_{da}}}

\newcommand{\z}{\ensuremath{\mathbf z}}

\setlist[itemize]{leftmargin=*}

\newlength\myindent
\DeclareMathOperator*{\argmin}{arg\,min}

\theoremstyle{plain}
\newtheorem{theorem}{Theorem}[section]
\newtheorem{corollary}[theorem]{Corollary}

\theoremstyle{definition}
\newtheorem{definition}{Definition}

\DeclarePairedDelimiter{\ceil}{\lceil}{\rceil}

\graphicspath{{figsallerton/}}

\title{\LARGE \bf
On the interplay of network structure and \\ gradient convergence in deep learning
}

\author{Vamsi K. Ithapu$^{\dagger}$ and Sathya N. Ravi$^{\dagger}$ and Vikas Singh$^{\ddagger,\dagger}$
\thanks{$^{\dagger}$Computer Sciences, University of Wisconsin-Madison}
\thanks{$^{\ddagger}$Biostatistics and Medical Informatics, University of Wisconsin-Madison}
}

\begin{document}

\maketitle
\thispagestyle{empty}
\pagestyle{empty}

\begin{abstract}

The regularization and output consistency behavior of dropout and layer-wise pretraining for learning deep networks have been fairly well studied.
However, our understanding of how the asymptotic convergence of backpropagation in deep architectures is related to the structural properties of the network 
and other design choices (like denoising and dropout rate) is less clear at this time. 
An interesting question one may ask is whether the network architecture and input data statistics may guide the choices of learning parameters and vice versa. 
In this work, we explore the association between such structural, distributional and learnability aspects vis-\`a-vis their interaction with parameter convergence rates. 
We present a framework to address these questions based on convergence of backpropagation for general nonconvex objectives using first-order information.
This analysis suggests an interesting relationship between feature denoising and dropout.
Building upon these results, we obtain a setup that provides systematic guidance regarding the choice of learning parameters and network sizes that 
achieve a certain level of convergence (in the optimization sense) often mediated by statistical attributes of the inputs.
Our results are supported by a set of experimental evaluations as well as independent empirical observations reported by other groups. 

\end{abstract}

%%%%%%%%%%%%%%%%%%%%%%%%%%%%%%%%%%%%%%%%%%%%%%%%%%%%%%%%%%%%%%%%%%%%%%%%%%%%%%%%%%%%%%%%%%%%%%%%%%%%%%%%%%%%%%%%%%%%%%%%%%%%%%%%%%%%%%%%%%%%%%%%%%%%
%%%%%%%%%%%%%%%%%%%%%%%%%%%%%%%%%%%%%%%%%%%%%%%%%%%%%%%%%%%%%%%%%%%%%%%%%%%%%%%%%%%%%%%%%%%%%%%%%%%%%%%%%%%%%%%%%%%%%%%%%%%%%%%%%%%%%%%%%%%%%%%%%%%%

\section{Introduction}
\label{sec:intro}

The successful deployment of deep learning in a broad spectrum of applications including localizing objects in images, analyzing particle accelerator data, converting speech into text, 
predicting the activity of drug molecules, and designing clinical trials 
\cite{lenz2015deep, krizhevsky2012imagenet, hinton2012deep, sutskever2014sequence, ithapu2015imaging} %mnih2013playing, szegedy2014going
provides compelling evidence that they can learn very complex concepts with fairly minimal feature engineering or preprocessing. 
This success is attributed to the idea of composing simple but non-linear modules that each transform the lower levels (raw or normalized data)
into a representation at a more abstract level \cite{hintonnature, bengio2009learning, bengio2013representation}.
While this high level representation learning procedure is 
quite general, the problem at hand may at least partly govern the choice of the architecture and require certain modifications to the algorithm. 
Motivated by various experimental considerations, several variants of deep architectures and corresponding regularization schemes have been developed 
\cite{lee2009convolutional, vincent2010stacked, baldi2014dropout, ioffe2015batch}. %goh2013top, wan2013regularization
Complementary to such algorithmic and empirical developments, 
there is also a growing recent interest in better understanding the mathematical properties of these algorithms. 
Over the last few years, several interesting results have been presented 
\cite{shao2011convergence, dauphin2014identifying, livni2014computational, bach2014breaking, arora2014provable, patel2015probabilistic, janzamin2015beating, hardt2015train}. %castillo2006very
While some of these address the hypothesis spaces and the sets of functions learnable by deep networks, the others analyze the parameter space learned via backprogapation.
A more detailed discussion about these works is included in Section \ref{sec:related}.

Our work is related to the foregoing recent results dealing with a theoretical characterization of 
the properties of deep networks. At the high level, we study the relationship between the learnability of the network (convergence of parameter estimation in an optimization sense)
and the structure of deep networks.
In particular, we are interested in a set of questions that can better elucidate the interplay between certain concepts that intuitively seem related, 
but for the most part, have been studied separately.  
For instance, what is the influence of the structural aspects of the architecture (number of layers, activation types) on practical 
considerations such as dropout rate, stepsizes, and so on? 
Can we {\it guide} the choice of some of these parameters based on their relationship to the others using information about the statistical moments of the data? 
More generally, can we evaluate which network converges faster, for a given learning problem? 
Or more specifically, what is the best network structure and activation functions for a given learning task?
Understanding such an interplay of the {\it architecture} and {\it data statistics}  vis-\`a-vis distinct {\it learning schemes} is necessary
to complement and even facilitate the continuing empirical successes of deep networks, especially in regimes where deep networks have not yet been thoroughly tested (like learning with small datasets).
Further, the interplay allows for relating the input data statistics to deep network learnability.
This may enable constructing nonlinearities beyond convolution and max-pooling (which remain important for computer vision datasets), for instance, to tackle 
structured data like brain images or genetic data where the sample size is a bottleneck.

{\it Overview:} The most commonly used procedure for parameter estimation in deep networks is mini-batch stochastic gradients  \cite{bottou2010large}. 
This involves deriving the average gradient from the error computed on a few training instances, adjusting the 
weights/parameters accordingly, and iterating until convergence. 
Our general goal is to tie this procedure (to the extent possible) to the network structure. 
In contrast to the other recent results \cite{janzamin2015beating}, 
we directly work with stochastic gradients with no strong assumptions on the data/network structure \cite{bach2014breaking}, 
which allows obtaining results that are quite generally applicable.
The starting point of our analysis is a recent work by \cite{ghadimi2013stochastic} dealing with 
the convergence of stochastic gradients for arbitrary nonconvex problems using a first-order oracle. 
We build upon and adapt this analysis by first addressing single-layer networks and unsupervised pretraining, and then, the more general case of multi-layer nets with dropout.
More importantly, apart from addressing the interplay of network structure and convergence, the algorithms we present, with minor tweaks
are easily deployable to the standard training pipeline.
Further, our bounds natively take into account the standard regularization schemes like dropout and layer-wise pretraining, making them potentially more useful in practice. 

A brief description of our {\bf contributions} are: 
{\bf (a)} Based on the idea of randomly stopping gradient updates from \cite{ghadimi2013stochastic}, we present a framework for analyzing 
mini-batch stochastic gradients on multi-layer deep networks, and prove gradient convergence of deep networks learned via dropout (with/without layer-wise pretraining).
{\bf (b)} Later, we derive explicit relationships between the network structure, learning parameters and input data statistics.
Our results corroborate many empirical studies, and may {\it guide} the choices of network/learning hyper-parameters. %(for modeling a given data corpus).  
%The analysis extends to convolutional and recurrent networks but these results are lengthy and omitted from this paper to keep the presentation succinct.  

%%%%%%%%%%%%%%%%%%%%%%%%%%%%%%%%%%%%%%%%%%%%%%%%%%%%%%%%%%%%%%%%%%%%%%%%%%%%%%%%%%%%%%%%%%%%%%%%%%%%%%%%%%%%%%%%%%%%%%%%%%%%%%%%%%%%%%%%%%%%%%%%%%%%%%%%%%%%%
%%%%%%%%%%%%%%%%%%%%%%%%%%%%%%%%%%%%%%%%%%%%%%%%%%%%%%%%%%%%%%%%%%%%%%%%%%%%%%%%%%%%%%%%%%%%%%%%%%%%%%%%%%%%%%%%%%%%%%%%%%%%%%%%%%%%%%%%%%%%%%%%%%%%%%%%%%%%%

\subsection{Related Work}\label{sec:related}
 
The body of literature addressing backpropagation in neural networks, and deep networks in general, is very large and dates back at least to the early $1970$s. 
Hence, we restrict the discussion only to those works that fall immediately within the context of this paper. 
While there are a number of seminal papers addressing variants of backpropagation, and stochastic gradients, focusing on the convergence of neural networks training
\cite{becker1988improving, magoulas1999improving}, a number of recent works %lecun1998gradient, vogl1988accelerating
\cite{ngiam2011optimization, bengio2012practical, dauphin2014identifying, janzamin2015beating, hardt2015train} %lecun2012efficient, andrychowicz2016learning
provide a fresh treatment of these problems and analyze efficient learning schemes in the context of deep networks specifically.
The solution space of backpropagation in this setting has also been addressed in recent results \cite{castillo2006very, shao2011convergence}, 
providing new insights into the types of functions learnable by deep networks \cite{livni2014computational, bach2014breaking}.
\cite{hardt2015train} have shown that better generalization can be achieved by ensuring smaller training times, 
while \cite{dauphin2014identifying} describes in detail the non-convex landscape of deep learning objectives and the goodness of local optima.
Beyond these optimization related works, several authors have independently addressed the regularization and learnability aspects of deep networks.
For example, \cite{wager2013dropout, baldi2014dropout} extensively analyzed dropout, \cite{patel2015probabilistic} develops a probabilistic theory of deep learning, 
and \cite{arora2014provable} studies the existence of deep representations. %by exploiting the structure of the network, 
\cite{arora2015deep} presents a generative model for ReLU-type deep networks, and very recently, \cite{wei2016network} studied the equivalence of arbitrary deep networks. 
Complimentary to these, \cite{janzamin2015beating} uses a tensor decomposition perspective.

%%%%%%%%%%%%%%%%%%%%%%%%%%%%%%%%%%%%%%%%%%%%%%%%%%%%%%%%%%%%%%%%%%%%%%%%%%%%%%%%%%%%%%%%%%%%%%%%%%%%%%%%%%%%%%%%%%%%%%%%%%%%%%%%%%%%%%%%%%%%%%%%%%%%%%%%%%%%%
%%%%%%%%%%%%%%%%%%%%%%%%%%%%%%%%%%%%%%%%%%%%%%%%%%%%%%%%%%%%%%%%%%%%%%%%%%%%%%%%%%%%%%%%%%%%%%%%%%%%%%%%%%%%%%%%%%%%%%%%%%%%%%%%%%%%%%%%%%%%%%%%%%%%%%%%%%%%%

\section{Preliminaries}
\label{sec:prelim}

\subsection{Notation}
\label{sec:notation}

Let $\x \in \mathbb{R}^{d_x}$ and $\y \in \mathbb{R}^{d_y}$ denote the input feature vector and the corresponding output (or label) respectively. 
Given multiple $\{\x,\y\} \in \mathcal{X}$, the unknown input-to-output mapping is modeled by a $L$-layered neural network ($L$-NN), 
which comprises of the input (or visible) unit $\x$, followed by $L-1$ hidden representations $\h^1,\ldots,\h^{L-1}$ and the output (or final) unit $\y$ \cite{bengio2009learning}.
The lengths of these $L+1$ representations are $d_0 = d_x, d_1,\ldots,d_{L-1}, d_L = d_y$ respectively.
Each layer transforms the representations from the previous layer by first applying an affine transform, followed by a non-linearity 
(in general, non-convex and not necessarily point-wise) \cite{bengio2009learning, bengio2013representation}.
The layer-wise transformations are denoted by $\W_l \in \mathbb{R}^{d_{l} \times d_{l-1}}$ for $l = 1,\ldots,L$.
Using these, the hidden representations are given by $\h^l = \sigma(\W_l,\h^{l-1})$ for $l=1,\ldots,L-1$ ($\h^0 = \x$), and the output layer is $\y = \sigma(\W_L,\h^{L-1})$, 
where $\sigma(\cdot)$ represents the non-linear function/mapping between layers. 
For a $1$-NN with no hidden layers, we simply have $\y = \sigma(\W,\x)$ where $\W$'s are the unknowns.
Note that the bias in the affine transformation is handled by augmenting features with $1$ whenever necessary.
We will restrict ourselves to point-wise sigmoid non-linearity i.e., for any ${\bf v} \in \mathbb{R}^d$, $\sigma({\bf v}) = \{\frac{1}{1+exp(-v_i)}\}$ ($i = 1,\ldots,d$).
The distributional hyper-parameters of interest are $\mu_\x = \frac{1}{d_x}\sum_j \mathbb{E}x_j$ and $\tau_\x = \frac{1}{d_x}\sum_j \mathbb{E}^2x_j$, 
which correspond to the average first moment and average squared first moment of the inputs respectively (the average is across the $d_x$ dimensions).
For simplicity we assume $\x\in[0,1]^{d_x}$ and $\y\in[0,1]^{d_y}$, and so $\mu_\x \in [0,1]$ and $\tau_\x \in [0,1]$. 

%%%%%%%%%%%%%%%%%%%%%%%%%%%%%%%%%%%%%%%%%%%%%%%%%%%%%%%%%%%%%%%%%%%%%%%%%%%%%%%%%%%%%%%%%%%%%%%%%%%%%%%%%%%%%%%%%%%%%%%%%%%%%%%%%%%%%%%%%%%%%%%%%%%%%
%%%%%%%%%%%%%%%%%%%%%%%%%%%%%%%%%%%%%%%%%%%%%%%%%%%%%%%%%%%%%%%%%%%%%%%%%%%%%%%%%%%%%%%%%%%%%%%%%%%%%%%%%%%%%%%%%%%%%%%%%%%%%%%%%%%%%%%%%%%%%%%%%%%%%

Consider the following minimization performed via mini-batch stochastic gradients \cite{bottou2010large},
\begin{equation}\label{eq:minrsg} \min_{\W} \quad f(\W) := \mathbb{E}_{\x,\y} \mathcal{L}(\x,\y;\W) \end{equation}
where $\mathcal{L}(\cdot)$ denotes some loss function parameterized by $\W$ and applied to data instances $\{\x,\y\}$.
Denote $\eta := \{\x,\y\} \sim \mathcal{X}$. The mini-batch stochastic gradient update using $B$ samples $\eta^1,\ldots,\eta^B$ is 
$\W \leftarrow \W - \gamma G(\eta;\W)$ where $\gamma$ is the stepsize. %\begin{equation}\label{eq:updatestep} \end{equation}
$G(\eta;\W)$ computed at $\W$ using the sample set $\eta^1,\ldots,\eta^B$ is given by
\begin{equation} \label{eq:update}
G(\eta;\W) = \frac{1}{B} \sum_{i=1}^B \nabla_{\W} \mathcal{L}(\eta^i;\W)
\end{equation}
Depending on $\mathcal{L}(\cdot)$, \eqref{eq:minrsg} corresponds to the backpropagation learning of different classes of neural networks using stochastic gradients. 
To address many such broad families, we exhaustively develop the analysis for three interesting and general classes of deep networks,
starting with single-layer networks, followed by unsupervised pretraining via box-constrained denoising autoencoders \cite{vincent2010stacked}, 
and finally multi-layer deep networks with dropout \cite{srivastava2014dropout}. 
Due to space restrictions, extensions to the more complex convolutional and recurrent networks including, fancier non-linearities like 
rectified linear units or max-outs \cite{goodfellow2013maxout}, are deferred to the longer version of this paper.
For each of these settings, the loss function $\mathcal{L}(\cdot)$ is defined as follows.
Due to space constraints we restrict the description of these classes to minimum, and refer the readers to \cite{bengio2009learning, vincent2010stacked, srivastava2014dropout}, 
and the references there in, for further and extensive details. 

%%%%%%%%%%%%%%%%%%%%%%%%%%%%%%%%%%%%%%%%%%%%%%%%%%%%%%%%%%%%%%%%%%%%%%%%%%%%%%%%%%%%%%%%%%%%%%%%%%%%%%%%%%%%%%%%%%%%%%%%%%%%%%%%%%%%%%%%%%%%%%%%%%%%%
%%%%%%%%%%%%%%%%%%%%%%%%%%%%%%%%%%%%%%%%%%%%%%%%%%%%%%%%%%%%%%%%%%%%%%%%%%%%%%%%%%%%%%%%%%%%%%%%%%%%%%%%%%%%%%%%%%%%%%%%%%%%%%%%%%%%%%%%%%%%%%%%%%%%%

\begin{itemize}
\item {\bf $1$-NN}: Single-layer Network 
\begin{equation}\label{eq:loss-singlenn} \mathcal{L}(\x,\y;\W) = \| \y - \sigma(\W\x) \|^2 \end{equation}
\item {\bf DA}: Box-constrained Denoising Autoencoder 
\begin{equation}\label{eq:loss-da} \mathcal{L}(\x,\y;\W) = \| \x - \sigma(\W^T\sigma(\W(\x * \z))) \|^2 \end{equation}
where $\W \in [-w_m,w_m]^{d_h \times d_v}$. $*$ denotes element-wise product and $\z$ is a binary vector of length $d_x$.
Given the denoising rate $\zeta$, the binary scalars $z_1,\ldots,z_{d_x} \sim Bernoulli(\zeta)$ are noise-indicators,
such that, whenever $z_i = 0$, the $i^{th}$ element of the input $\x$, $x_i$ is nullified. We denote $\x * \z$ as $\tx$. 
$[-w_m,w_m]$ is the box-constraint on the unknowns $\W$, which forces the learned representations to {\em not} saturate around $0$ and/or $1$, 
and has been widely used in variants of both autoencoder design and backpropagation itself \cite{vincent2010stacked, erhan2010does}. 
The nature of this constraint is similar in part to other regularization schemes like \cite{bengio2012practical} and \cite{ngiam2011optimization, srivastava2014dropout}.
%Although feature dropout based fully-supervised models have been shown to achieve good generalization accuracy, 
%unsupervised {\it pretraining} was important for the initial revival of deep networks \cite{bengio2009learning, erhan2010does}. 
%Hence, we analyze DA to ensure that we cover this classical deep learning in our analysis.

%%%%%%%%%%%%%%%%%%%%%%%%%%%%%%%%%%%%%%%%%%%%%%%%%%%%%%%%%%%%%%%%%%%%%%%%%%%%%%%%%%%%%%%%%%%%%%%%%%%%%%%%%%%%%%%%%%%%%%%%%%%%%%%%%%%%%%%%%%%%%%%%%%%%%
%%%%%%%%%%%%%%%%%%%%%%%%%%%%%%%%%%%%%%%%%%%%%%%%%%%%%%%%%%%%%%%%%%%%%%%%%%%%%%%%%%%%%%%%%%%%%%%%%%%%%%%%%%%%%%%%%%%%%%%%%%%%%%%%%%%%%%%%%%%%%%%%%%%%%

\item {\bf $L$-NN}: Multi-layer Network 
\begin{equation}\label{eq:loss-mulnn} \begin{aligned}
\h^0 = \x &;\quad \h^l = \sigma \left( \W_l (\h^{l-1} * \z^l) \right) \\
\mathcal{L}(\x,\y;\W) &= \| \y - \sigma \left( \W_L (\h^{L-1} * \z^L) \right) \|^2
\end{aligned} \end{equation}
where $l = 1,\ldots,L-1$. Recall the dropout scheme in training deep networks which address the overfitting problem \cite{srivastava2014dropout}.
In each gradient update iteration, a random fraction of the hidden and/or input units are dropped based on the (given) dropout rates $\zeta_1,\ldots,\zeta_L$ \cite{srivastava2014dropout}. 
Similar to DA, the dropped out units for each layer are denoted by a set of binary vectors $\z^1,\ldots,\z^L$ such that $z^l_i \sim Bernoulli(\zeta_l)$ for $l=1,\ldots,L$.
Within each iteration, this results in randomly sampling a smaller sub-network with approximately $\prod_{l=1}^L \zeta_l$ fraction of all the transformation parameters.
Only these $\prod_{l=1}^L \zeta_l$ fraction of weights are updated in the current iteration, while the rest are not, and the re-sampling and updating process is repeated. 
For simplicity of analysis we use the same dropout rate, denoted by $\zeta$, for all layers. %(note the abuse of notation here with respect to denoising rate in DA) 
With some abuse of notation, we use $\W^{k}_{l}$ to represent the $k^{th}$ gradient update of the $l^{th}$ layer transformation $\W_l$. 
\begin{itemize}
\item A $L$-NN may be pretrained layer-wise before supervised tuning \cite{bengio2009learning, vincent2010stacked, erhan2010does}.
Here, the $L-1$ hidden layers are first pretrained, for instance, using the box-constrained DA from \eqref{eq:loss-da}. 
This gives the estimates of the $L-1$ transformations $\W_1,\ldots,\W_{L-1}$, which (along with $\y$s) are then used to {\it initialize} the $L$-NN.
This `pretrained' $L$-NN is then learned using dropout as in \eqref{eq:loss-mulnn}.
This is the classical deep learning regime \cite{bengio2009learning, erhan2010does}, and we discuss this case separately from the fully-supervised version which has {\it no} layer-wise pretraining.
Several studies have already shown interesting empirical relationships between dropout and the DA \cite{wager2013dropout}, 
and we complement this body of work by providing explicit relationships between them. 
\end{itemize}
\end{itemize}

%%%%%%%%%%%%%%%%%%%%%%%%%%%%%%%%%%%%%%%%%%%%%%%%%%%%%%%%%%%%%%%%%%%%%%%%%%%%%%%%%%%%%%%%%%%%%%%%%%%%%%%%%%%%%%%%%%%%%%%%%%%%%%%%%%%%%%%%%%%%%%%%%%%%%
%%%%%%%%%%%%%%%%%%%%%%%%%%%%%%%%%%%%%%%%%%%%%%%%%%%%%%%%%%%%%%%%%%%%%%%%%%%%%%%%%%%%%%%%%%%%%%%%%%%%%%%%%%%%%%%%%%%%%%%%%%%%%%%%%%%%%%%%%%%%%%%%%%%%%

\subsection{Roadmap}\label{sec:roadmap} 

The gradient update in \eqref{eq:update} is central to the ideas described in this paper. 
By {\it tracking} the behavior of these gradients as they propagate across multiple layers of the network, with minimal assumptions on the loss function $\mathcal{L}(\eta;\W)$, 
we can assess the convergence of the overall parameter estimation scheme while taking into account the influence (and structure) of each of the layers involved.
Since the updates are stochastic, ideally, we are interested in the ``expected'' gradients over a certain number of iterations, fixed ahead of time. 
Motivated by this intuition, our main idea adapted from \cite{ghadimi2013stochastic}, is to randomly sample the number of gradient update iterations. 
Specifically, let $N$ denote the maximum possible number of iterations that can be performed keeping in mind the memory and time constraints (in general, $N$ is very large). 
The stopping distribution $\mathbb{P}_R(\cdot)$ gives the probability that $k^{th}$ iteration ($k = 1,\ldots,N$) is the {\it last or stopping} iteration. 
We denote this randomly sampled stopping iteration as $R \in \{1,\ldots,N\}$, and so,
\begin{equation}\label{eq:stopprob}
\mathbb{P}_R(\cdot) := \frac{p^k_R}{\sum_{k=1}^N p^k_R} \quad\text{where}\quad p^k_R = Pr(R = k)
\end{equation}
$\mathbb{P}_R(\cdot)$ can either be fixed ahead of time or learned from a hyper-training procedure. 
An alternate way to interpret the stopping distribution is by observing that $p^k_R$ represents the probability that the estimates ($\W$'s) 
at the $k^{th}$ iteration are desired final solutions returned by the learning procedure. %(This aspect will be used in the deriving the results). 

%%%%%%%%%%%%%%%%%%%%%%%%%%%%%%%%%%%%%%%%%%%%%%%%%%%%%%%%%%%%%%%%%%%%%%%%%%%%%%%%%%%%%%%%%%%%%%%%%%%%%%%%%%%%%%%%%%%%%%%%%%%%%%%%%%%%%%%%%%%%%%%%%%%%%
%%%%%%%%%%%%%%%%%%%%%%%%%%%%%%%%%%%%%%%%%%%%%%%%%%%%%%%%%%%%%%%%%%%%%%%%%%%%%%%%%%%%%%%%%%%%%%%%%%%%%%%%%%%%%%%%%%%%%%%%%%%%%%%%%%%%%%%%%%%%%%%%%%%%%

By proceeding with this {\it random stopping mini-batch stochastic gradients}, 
and using some Lipschitz properties of the objective and certain distributional characteristics of the input data, 
we can analyze the three loss functions \eqref{eq:loss-singlenn}, \eqref{eq:loss-da} and \eqref{eq:loss-mulnn}.
The stopping iteration $R$ is random and the loss function in \eqref{eq:minrsg} includes an expectation over data instances $\eta = \{\x,\y\}$.
Therefore, we are interested in the {\it expectation} of the gradients $\nabla_{\W} f(\W^k)$ computed over $R \sim \mathbb{P}_R(\cdot)$ and $\eta \sim \mathcal{X}$.
Few of the hyper-parameters that play significant role are, 
the variance of $G(\eta;\W)$ in \eqref{eq:update} denoted by $e^{s}$, $e^{da}$ and $e^{m}$ for single-layer, pretraining and multi-layer cases respectively, 
the distributional hyper-parameters $\mu_x$ and $\tau_x$, and the denoising/dropout rate $\zeta$ along with the box-constraint.
%This seemingly small variation to the standard backpropagation leads to gradient convergence bounds that incorporate the network depth and lengths, 
%and other learning choices like dropout or denoising rate in a very natural manner. 
%
%%%%%%%%%%%%%%%%%%%%%%%%%%%%%%%%%%%%%%%%%%%%%%%%%%%%%%%%%%%%%%%%%%%%%%%%%%%%%%%%%%%%%%%%%%%%%%%%%%%%%%%%%%%%%%%%%%%%%%%%%%%%%%%%%%%%%%%%%%%%%%%%%%%%%
%%%%%%%%%%%%%%%%%%%%%%%%%%%%%%%%%%%%%%%%%%%%%%%%%%%%%%%%%%%%%%%%%%%%%%%%%%%%%%%%%%%%%%%%%%%%%%%%%%%%%%%%%%%%%%%%%%%%%%%%%%%%%%%%%%%%%%%%%%%%%%%%%%%%%
%
Apart from the convergence bounds, we also derive sample sizes required for large deviation estimates of $\W$'s which marginalizes the influence of the random stopping iteration.
This leads to bounds on training time and the minimum required training dataset size, which in turn can be used to ensure certain level of generalization using existing results \cite{hardt2015train}.
The treatment for the simple case of single-layer networks serves as basis for the more general settings.
Due to space constraints, the proofs for all the technical results are included in an adjoining technical report at 
{\footnotesize \url{http://pages.cs.wisc.edu/\%7Evamsi/files/techrep-dlinterplay.pdf}} 

%%%%%%%%%%%%%%%%%%%%%%%%%%%%%%%%%%%%%%%%%%%%%%%%%%%%%%%%%%%%%%%%%%%%%%%%%%%%%%%%%%%%%%%%%%%%%%%%%%%%%%%%%%%%%%%%%%%%%%%%%%%%%%%%%%%%%%%%%%%%%%%%%%%%%
%%%%%%%%%%%%%%%%%%%%%%%%%%%%%%%%%%%%%%%%%%%%%%%%%%%%%%%%%%%%%%%%%%%%%%%%%%%%%%%%%%%%%%%%%%%%%%%%%%%%%%%%%%%%%%%%%%%%%%%%%%%%%%%%%%%%%%%%%%%%%%%%%%%%%

{\bf Why gradient norm?} 
We may ask if characterizing the behavior of the gradients is ideal for the goal of characterizing the interplay. 
There are more than a few reasons why this strategy is at least sensible. 
First, note that it is NP-hard to check local minima even for simple non-convex problems \cite{murty1987some} --- 
so, an analysis using the norms of the gradients is an attractive alternative, especially, if it leads to a similar main result. 
Second, a direct way of approaching our central question of the architecture and convergence interplay is by analyzing the gradients themselves.
Clearly, learnability of the network would be governed by the goodness of the estimates, which in turn depend on the convergence of stochastic gradients. 
In most cases, this naturally requires an asymptotic analysis of the norm, which, 
for nonconvex objectives with minimal assumptions (like Lipschitz continuity) was unavailable until very recently \cite{ghadimi2013stochastic}.
Third, working with the gradient norm directly allows for assessing faster training times, which, as argued in \cite{hardt2015train}, is vital for better generalization. 
Alternatively, using the gradient convergence as a surrogate for this training time allows for modulating the network structure to improve generalization.

%%%%%%%%%%%%%%%%%%%%%%%%%%%%%%%%%%%%%%%%%%%%%%%%%%%%%%%%%%%%%%%%%%%%%%%%%%%%%%%%%%%%%%%%%%%%%%%%%%%%%%%%%%%%%%%%%%%%%%%%%%%%%%%%%%%%%%%%%%%%%%%%%%%%%
%%%%%%%%%%%%%%%%%%%%%%%%%%%%%%%%%%%%%%%%%%%%%%%%%%%%%%%%%%%%%%%%%%%%%%%%%%%%%%%%%%%%%%%%%%%%%%%%%%%%%%%%%%%%%%%%%%%%%%%%%%%%%%%%%%%%%%%%%%%%%%%%%%%%%

\section{Single-layer Networks}
\label{sec:single}

Consider a $1$-NN with the corresponding loss function from \eqref{eq:loss-singlenn}.
We learn it via the random stopping mini-batch stochastic gradients (batch size $B$ and $R \sim \mathbb{P}_R(k)$, $k = 1,\ldots,N$).  
This learning procedure is summarized in Alg. 1 in the supplemental technical report (see the link from Section \ref{sec:prelim}); 
referred from here on as single-layer randomized stochastic gradients, {\it single-layer RSG}. %$*$ in Alg. \ref{alg:single-nn} refers to element-wise product.
$\W^1$ and $\W^R$ are the initial and final estimates respectively, and $\gamma^k$ is the stepsize at the $k^{th}$ iteration.
With no prior information about how to setup the stopping distribution, one would simply choose $\mathbb{P}_R(k) := Unif[1,N]$.
The first result summarizes the decay of the expected gradients in single-layer RSG for this setting. 
$D_f = f(\W^1) - f^{*}$ is the initial deviation of the objective from unknown optimum $\W^{*}$.
\begin{theorem}[{\bf $1$-NN, Constant Stepsize}] \label{thm:expgrad}
Consider a single-layer RSG with {\it no} dropout and constant stepsize $\gamma^k = \gamma \forall k$.
Let $e^s_\gamma = (1 - \frac{13}{16}\gamma)$, $e^s = \frac{13d_xd_y}{256}$ and $R \sim Unif[1,N]$. The expected gradients are given by
\begin{equation} \label{eq:conv1nn}
\mathbb{E}_{R,\eta} (\| \nabla_{\W} f(\W^R)\|^2) \leq \frac{1}{e^s_\gamma} \left( \frac{D_f}{N\gamma} + \frac{e^s\gamma}{B} \right) 
\end{equation}
and the optimal constant stepsize is $\gamma_o = \sqrt{\frac{Bf(\W^1)}{e^sN}}$
\end{theorem}

%%%%%%%%%%%%%%%%%%%%%%%%%%%%%%%%%%%%%%%%%%%%%%%%%%%%%%%%%%%%%%%%%%%%%%%%%%%%%%%%%%%%%%%%%%%%%%%%%%%%%%%%%%%%%%%%%%%%%%%%%%%%%%%%%%%%%%%%%%%%%%%%%%%%%%%%%%%%
%%%%%%%%%%%%%%%%%%%%%%%%%%%%%%%%%%%%%%%%%%%%%%%%%%%%%%%%%%%%%%%%%%%%%%%%%%%%%%%%%%%%%%%%%%%%%%%%%%%%%%%%%%%%%%%%%%%%%%%%%%%%%%%%%%%%%%%%%%%%%%%%%%%%%%%%%%%% 

{\it Remarks:} 
We should point out that the {\em asymptotic} behavior of gradients in backpropagation, including variants with adaptive stepsizes, momentum etc. 
have been well studied \cite{magoulas1999improving, becker1988improving, castillo2006very, shao2011convergence}. 
This aspect is {\it not} novel to our work specifically, as also described in Section \ref{sec:intro}.
However, to our knowledge, relatively few `explicit' results about the convergence {\it rates} are known; 
although imposing restrictions on the objective does lead to improved guarantees \cite{ahmad1990asymptotic}.
This may be because of the lack of such results in the numerical optimization literature in the context of general nonconvex objectives.
The recent result in \cite{ghadimi2013stochastic} presents one of the first such results addressing general nonconvex objectives with a first-order oracle. 
Consequently, we believe that Theorem \ref{thm:expgrad} gives non-trivial information and, as we will show in later sections, leads to interesting new 
results in the context of neural networks.
%This result serves as a building blocks for analyzing feature denoising and multi-layer dropout later in Section \ref{sec:pretrain} and \ref{sec:multi}.  
We now explain \eqref{eq:conv1nn} briefly.

%%%%%%%%%%%%%%%%%%%%%%%%%%%%%%%%%%%%%%%%%%%%%%%%%%%%%%%%%%%%%%%%%%%%%%%%%%%%%%%%%%%%%%%%%%%%%%%%%%%%%%%%%%%%%%%%%%%%%%%%%%%%%%%%%%%%%%%%%%%%%%%%%%%%%%%%%%%%
%%%%%%%%%%%%%%%%%%%%%%%%%%%%%%%%%%%%%%%%%%%%%%%%%%%%%%%%%%%%%%%%%%%%%%%%%%%%%%%%%%%%%%%%%%%%%%%%%%%%%%%%%%%%%%%%%%%%%%%%%%%%%%%%%%%%%%%%%%%%%%%%%%%%%%%%%%%%

While the first term corresponds to the goodness of fit of the network (showing the influence of $D_f$), 
the second term encodes the network degrees of freedom ($e^s$), and is larger for big (or fatter) networks.
Clearly, the bounds decreases as $N$ (and/or $B$) increase, and it is interesting to see that the effect of network size ($d_xd_y$) is negligible for large batch sizes.
The second term in \eqref{eq:conv1nn} induces a kind of `bias', that depends on $e^s$ and $B$.
As $B$ increases, \eqref{eq:update} will be much closer to a full-batch gradient update, there by reducing this bias 
(a classical property of mini-batch stochastic gradients \cite{lecun2012efficient, bengio2012practical}). 
Further, this term indicates that larger batch sizes are necessary for fatter networks with large $d_x$ or $d_y$. 
One caveat of the generality of \eqref{eq:conv1nn} is that it might be loose in the worst case where $D_f \sim 0$ 
(i.e., when $\W^1$ is already a good estimate of the stationary point). 
$e^s_\gamma$ increases the overall bound as $\gamma$ increases.
The optimal stepsize $\gamma_o$ in Theorem \ref{thm:expgrad} is calculated by balancing the two terms in \eqref{eq:conv1nn}, 
using which we see that the expected gradients are $\mathcal{O}\left(\frac{1}{\sqrt{N}}\right)$ for a given network and $\mathcal{O}(\sqrt{d_xd_y})$ for a given N.

%%%%%%%%%%%%%%%%%%%%%%%%%%%%%%%%%%%%%%%%%%%%%%%%%%%%%%%%%%%%%%%%%%%%%%%%%%%%%%%%%%%%%%%%%%%%%%%%%%%%%%%%%%%%%%%%%%%%%%%%%%%%%%%%%%%%%%%%%%%%%%%%%%%%%%%%%%%%
%%%%%%%%%%%%%%%%%%%%%%%%%%%%%%%%%%%%%%%%%%%%%%%%%%%%%%%%%%%%%%%%%%%%%%%%%%%%%%%%%%%%%%%%%%%%%%%%%%%%%%%%%%%%%%%%%%%%%%%%%%%%%%%%%%%%%%%%%%%%%%%%%%%%%%%%%%%%

We see from the proof of Theorem \ref{thm:expgrad} that (see technical report), 
allowing constant $\gamma^k$s directly corresponds to choosing uniform $\mathbb{P}_R(\cdot)$ i.e., they are {\it equivalent} in some sense. 
Nevertheless, using alternate $\mathbb{P}_R(\cdot)$ in tandem with constant $\gamma^k$s results in different sets of constants in \eqref{eq:conv1nn}, 
while the overall (high-level) dependence on $N$ and network size $d_xd_y$ would remain the same. 
Beyond constant $\gamma^k$s, the most common setting is to use decaying $\gamma^k$s.
The corresponding result is in Theorem \ref{thm:expgrad-noncon} with $\gamma^k = \frac{\gamma}{k^{\rho}}$ (for some $\rho>0$), 
and $\mathcal{H}_N(\theta) = \sum_{i=1}^N \frac{1}{i^{\theta}}$ is the generalized harmonic number.
\begin{corollary}[{\bf $1$-NN, Decreasing Stepsize}] \label{thm:expgrad-noncon}
Consider a single-layer RSG with {\it no} dropout and stepsizes $\gamma^k = \frac{\gamma}{k^{\rho}}$. 
Let $e^s = \frac{13d_xd_y}{256}$ and the probability of stopping at $k^{th}$ iteration $p^k_R = \gamma^k (1 - \frac{13}{16}\gamma^k)$. 
The expected gradients are given by
\begin{equation} \label{eq:conv1nn-noncon}
\mathbb{E}_{R,\eta} (\| \nabla_{\W} f(\W^R)\|^2) \leq \frac{16}{3\mathcal{H}_N(\rho)} \left( \frac{D_f}{\gamma} + \frac{e^s\gamma\mathcal{H}_N(2\rho)}{B} \right) 
\end{equation} \end{corollary}

%%%%%%%%%%%%%%%%%%%%%%%%%%%%%%%%%%%%%%%%%%%%%%%%%%%%%%%%%%%%%%%%%%%%%%%%%%%%%%%%%%%%%%%%%%%%%%%%%%%%%%%%%%%%%%%%%%%%%%%%%%%%%%%%%%%%%%%%%%%%%%%%%%%%%%%%%%%%
%%%%%%%%%%%%%%%%%%%%%%%%%%%%%%%%%%%%%%%%%%%%%%%%%%%%%%%%%%%%%%%%%%%%%%%%%%%%%%%%%%%%%%%%%%%%%%%%%%%%%%%%%%%%%%%%%%%%%%%%%%%%%%%%%%%%%%%%%%%%%%%%%%%%%%%%%%%%

{\it Remarks:} 
Whenever $\rho \approx 0$, the stepsizes are approximately constant, and we have $\mathcal{H}_N(\rho) \approx N$ (assuming that $\gamma < 1$).
Here the bound in \eqref{eq:conv1nn-noncon} is at least as large as \eqref{eq:conv1nn} making the two results consistent with each other.
The dependence of $\rho$, and its interaction with $N$ and other hyper-parameters, on the decay of the expected gradients is reasonably complex.
Nevertheless, broadly, as $\rho$ increases, the bound first decreases and eventually increases becoming more looser. 
This trend is expected. To see this, first observe that large $\rho$ implies strong decay of the stepsizes.
For sufficiently small $\gamma$, this results in stopping iteration probability $p^k_R$  that decreases as $k$ increases (refer to its definition from Theorem \ref{thm:expgrad-noncon}), 
i.e., large $\rho$ results in $R \ll N$.
The bound is implying the trivial fact that the expected gradients are going to be large whenever the gradient updating is stopped early.  

%%%%%%%%%%%%%%%%%%%%%%%%%%%%%%%%%%%%%%%%%%%%%%%%%%%%%%%%%%%%%%%%%%%%%%%%%%%%%%%%%%%%%%%%%%%%%%%%%%%%%%%%%%%%%%%%%%%%%%%%%%%%%%%%%%%%%%%%%%%%%%%%%%%%%%%%%%%%
%%%%%%%%%%%%%%%%%%%%%%%%%%%%%%%%%%%%%%%%%%%%%%%%%%%%%%%%%%%%%%%%%%%%%%%%%%%%%%%%%%%%%%%%%%%%%%%%%%%%%%%%%%%%%%%%%%%%%%%%%%%%%%%%%%%%%%%%%%%%%%%%%%%%%%%%%%%%

Overall, these observations clearly imply that \eqref{eq:conv1nn} and \eqref{eq:conv1nn-noncon} 
are capturing all the intuitive trends one would expect to see from using stochastic gradients, in turn, making the analysis and results all the more useful.
Lastly, it may not be appropriate to bias $R$ to be far smaller than $N$. 
One can relax this and instead use arbitrary $\mathbb{P}_R(\cdot)$ that is monotonically {\it increasing}, i.e., $p^k_R \leq p^{k+1}_R \forall k$, there by pushing $R$ towards $N$ with high probability.
A similar decay of gradients (like Corollary \ref{thm:expgrad-noncon}) result can be obtained for this case, but we omit it from this shorter version of the paper. 
%
%%%%%%%%%%%%%%%%%%%%%%%%%%%%%%%%%%%%%%%%%%%%%%%%%%%%%%%%%%%%%%%%%%%%%%%%%%%%%%%%%%%%%%%%%%%%%%%%%%%%%%%%%%%%%%%%%%%%%%%%%%%%%%%%%%%%%%%%%%%%%%%%%%%%%%%%%%%%
%%%%%%%%%%%%%%%%%%%%%%%%%%%%%%%%%%%%%%%%%%%%%%%%%%%%%%%%%%%%%%%%%%%%%%%%%%%%%%%%%%%%%%%%%%%%%%%%%%%%%%%%%%%%%%%%%%%%%%%%%%%%%%%%%%%%%%%%%%%%%%%%%%%%%%%%%%%%
%
Theorem \ref{thm:expgrad} and Corollary \ref{thm:expgrad-noncon} describe convergence of $1$-NN for {\em one run} of stochastic gradient. 
In practice, one is more interested in a large deviation bound over multiple runs, especially because of the randomization over $\eta$ and $R$ (see \eqref{eq:conv1nn}).
We define such a large deviation estimate, using $\W^{R_1},\ldots,\W^{R_T}$ computed from $T>1$ {\it independent} runs of single-layer RSG, 
and compute the minimum $N$ required to achieve such an estimate. 
\begin{definition}[{\bf $(\epsilon,\delta)$-solution}]\label{thm:epsdel}
Given $\epsilon>0$ and $0<\delta\ll 1$, an $(\epsilon,\delta)$-solution of a single-layer network is given by $\argmin_t \|\nabla_{\W} f(\W^{R_t})\|^2$ such that 
$Pr\left( \min_t \| \nabla_{\W} f(\W^{R_t}) \|^2 \leq \epsilon \right) \geq 1-\delta$.
\end{definition}
\begin{corollary}[{\bf Computational Complexity}] \label{thm:conv1nn}
To compute a $(\epsilon,\delta)$-solution for single-layer RSG with {\it no} dropout, optimal constant stepsizes and $\mathbb{P}_R := Unif[1,N]$, we need 
\begin{equation}\label{eq:iter1nn}  
N(\epsilon,\delta) \geq \frac{4f(\W^1)e^s}{B\delta^{2/T}\epsilon^2} (1 + \bar{\delta})^2 \quad,\quad \bar{\delta} = \frac{13B\epsilon\delta^{1/T}}{32e^s} 
\end{equation} \end{corollary}
{\it Remarks:} 
To illustrate the practical usefulness of this result, consider an example network with $d_x=100$, $d_y=5$ and say $f(\W^1)\sim d_y$.
For a $(0.05,0.05)$-solution with $T=10$ runs and a batchsize $B=50$, \eqref{eq:iter1nn} gives $N>7.8\times 10^3$.
If the number of epochs is $C=200$, then under the (reasonable) assumption that $SC \approx BN$ (where $S$ is sample size), this leads to $\sim 2000$ instances. 
Clearly, \eqref{eq:iter1nn} implies that $N$ increases as the network size ($d_xd_y$) and $f(\W^1)$ increase, 
and for obtaining good large deviation estimates (i.e., $\epsilon$, $\delta \approx 0$), the number of required iterations is large.
Corollary \ref{thm:conv1nn} does not use any information about the input data statistics (e.g., moments), and so expectedly, \eqref{eq:iter1nn} may overestimate $N$ and $S$. 
Later, we make use of such data statistics to analyze networks with hidden layers, while the rest of the recipe extends from the $1$-NN setting.

%%%%%%%%%%%%%%%%%%%%%%%%%%%%%%%%%%%%%%%%%%%%%%%%%%%%%%%%%%%%%%%%%%%%%%%%%%%%%%%%%%%%%%%%%%%%%%%%%%%%%%%%%%%%%%%%%%%%%%%%%%%%%%%%%%%%%%%%%%%%%%%%%%%%%%%%%%%%
%%%%%%%%%%%%%%%%%%%%%%%%%%%%%%%%%%%%%%%%%%%%%%%%%%%%%%%%%%%%%%%%%%%%%%%%%%%%%%%%%%%%%%%%%%%%%%%%%%%%%%%%%%%%%%%%%%%%%%%%%%%%%%%%%%%%%%%%%%%%%%%%%%%%%%%%%%%%

{\it Choosing $\mathbb{P}_R(\cdot)$:} 
An important issue in the results presented is the choice of $\mathbb{P}_R(\cdot)$. 
Firstly, Theorem \ref{thm:expgrad} and Corollary \ref{thm:expgrad-noncon} cover the typical choices of stopping criteria.
Secondly, from a practical stand-point, once can use multiple choices of $\mathbb{P}_R(\cdot)$ from a dictionary of distributions $\mathcal{P}$, and choose the best one via some cross-validation. 
For instance, the best $\mathbb{P}_R(\cdot) \in \mathcal{P}$ can be selected based on a validation dataset 
either by directly computing the empirical average of the gradients, or using alternate measures like generalization performance. 
This is valid because several recent results justify using number of training iterations as a `surrogate' for the generalization performance \cite{hardt2015train}. 
Further, the analysis also allows for probing the nature of the $R$ after fixing $\mathbb{P}_R(\cdot)$ i.e., depending on the Hessian at $R^{th}$ iteration, 
gradient updates may be continued if needed, and the technical results presented here would still hold for this post-hoc increase in iterations.
The longer version of the paper will have extensive analysis about the influence and appropriate choices of $\mathbb{P}_R(\cdot)$

%%%%%%%%%%%%%%%%%%%%%%%%%%%%%%%%%%%%%%%%%%%%%%%%%%%%%%%%%%%%%%%%%%%%%%%%%%%%%%%%%%%%%%%%%%%%%%%%%%%%%%%%%%%%%%%%%%%%%%%%%%%%%%%%%%%%%%%%%%%%%%%%%%%%%%%%%%%%
%%%%%%%%%%%%%%%%%%%%%%%%%%%%%%%%%%%%%%%%%%%%%%%%%%%%%%%%%%%%%%%%%%%%%%%%%%%%%%%%%%%%%%%%%%%%%%%%%%%%%%%%%%%%%%%%%%%%%%%%%%%%%%%%%%%%%%%%%%%%%%%%%%%%%%%%%%%%

\section{Unsupervised Pretraining}
\label{sec:pretrain}

Building upon the results from \ref{thm:expgrad}--\ref{thm:conv1nn}, we now consider single-layer networks that perform unsupervised learning 
via box-constrained DA \cite{vincent2010stacked} (see the loss function from \eqref{eq:loss-da}).
Unlike the single-layer setting, because of the constraint on $\W$, we will now be interested in the expected {\it projected} gradients $\nabla_{\W} \tf(\W)$, 
which simply are the Euclidean projections of $\nabla_{\W} f(\W)$ on $\W \in [-w_m,w_m]^{d_h \times d_x}$.
Alg. 2 in the supplemental technical report summaries this DA RSG procedure.
%$P_{\W}$ denotes the projection operator onto the constraint set $[-w_m,w_m]^{d_h \times d_x}$.
To ensure broader discussion of the interplay (of network structure and learning) we restrict ourselves to the case of constant $\gamma^k$s with $R \sim Unif[1,N]$.
We point out that the trends inferred from our results would still be appropriate for alternative stepsizes and $\mathbb{P}_R(\cdot)$.
The following result bounds $\nabla_{\W} \tf(\W)$ for DA RSG. 
$e^{da}$ and $e^{da}_\gamma$ encode the network structure and learning constants.
\begin{theorem}[{\bf DA, constant stepsize}] \label{thm:expgradda}
Consider a DA RSG with $\W \in [-w_m,w_m]^{d_h\times d_x}$. Let
\begin{equation} \label{eq:convdaparam-eda}
e^{da} = \frac{d_xd_h}{16} \left[ 1 + \frac{\zeta d_xw_m}{4}\mu_{\x} + \left(\frac{5\zeta}{16} - \frac{\zeta^2}{4}\right)(\zeta d_xw_m)^2 \tau_{\x} \right] 
\end{equation}
and $\uda$ denote the Lipschitz constant of $\nabla_{\W} f(\W)$ with loss function from \eqref{eq:loss-da}.
Using constant stepsize $\gamma<\frac{2}{\uda}$ and denoting $e^{da}_\gamma = 1 - \frac{\uda}{2}\gamma$, the expected projected gradients are
\begin{equation} \label{eq:convda}
\mathbb{E} (\| \nabla_{\W} \tf(\W^R)\|^2) \leq \frac{D_f}{N\gamma e^{da}_\gamma} + \frac{e^{da}}{B} \left( 1+\frac{1}{e^{da}_\gamma}\right) 
\end{equation}
and the optimal stepsize is $\gamma_o \approx \sqrt{\frac{2BD_f}{\uda e^{da}N}}$
\end{theorem}

%%%%%%%%%%%%%%%%%%%%%%%%%%%%%%%%%%%%%%%%%%%%%%%%%%%%%%%%%%%%%%%%%%%%%%%%%%%%%%%%%%%%%%%%%%%%%%%%%%%%%%%%%%%%%%%%%%%%%%%%%%%%%%%%%%%%%%%%%%%%%%%%%%%%%
%%%%%%%%%%%%%%%%%%%%%%%%%%%%%%%%%%%%%%%%%%%%%%%%%%%%%%%%%%%%%%%%%%%%%%%%%%%%%%%%%%%%%%%%%%%%%%%%%%%%%%%%%%%%%%%%%%%%%%%%%%%%%%%%%%%%%%%%%%%%%%%%%%%%%

{\it Remarks:} 
Similar to \eqref{eq:conv1nn} from Theorem \ref{thm:expgrad} for $1$-NN, the above result in \eqref{eq:convda} combines the contributions from 
the output goodness of fit ($D_f$), the number of free parameters ($d_hd_x$) and the stepsize choice ($\gamma$, $e^{da}_\gamma$).
All the remarks from Theorem \ref{thm:expgrad} would still apply here -- $D_f$ is balanced out by the number of iterations $N$, 
the second term involving the variance of gradients ($e^{da}$) and the batchsize controls the `bias' from the network's degrees of freedom. 
However, unlike $1$-NN, here the dependence on the network structure is much more involved. 
The input and hidden layer lengths do not contribute equally (see $e^{da}$ from \eqref{eq:convdaparam-eda}) which can be partly explained 
by the asymmetric structure of the loss (refer to \eqref{eq:loss-da}).
For smaller constraint sets i.e., small $w_m$, and hence small $e^{da}$, we expect the projected gradients to typically have small magnitude, which is clearly implied by \eqref{eq:convda}.
In practice, $w_m$ is reasonably small \cite{pathak2015constrained} and, in general, 
$W_{ij}$'s have been shown to emulate a `fat' Gaussian with mean centered around zero \cite{bellido1993backpropagation}. %, blundell2015weight}. 

%%%%%%%%%%%%%%%%%%%%%%%%%%%%%%%%%%%%%%%%%%%%%%%%%%%%%%%%%%%%%%%%%%%%%%%%%%%%%%%%%%%%%%%%%%%%%%%%%%%%%%%%%%%%%%%%%%%%%%%%%%%%%%%%%%%%%%%%%%%%%%%%%%%%%
%%%%%%%%%%%%%%%%%%%%%%%%%%%%%%%%%%%%%%%%%%%%%%%%%%%%%%%%%%%%%%%%%%%%%%%%%%%%%%%%%%%%%%%%%%%%%%%%%%%%%%%%%%%%%%%%%%%%%%%%%%%%%%%%%%%%%%%%%%%%%%%%%%%%%

{\bf Denoising Rates versus Network structure:} 
\eqref{eq:convdaparam-eda} and its resulting structure in \eqref{eq:convda} seem to imply a non-trivial interplay between the network size $d_hd_x$, 
the data statistics (via $\mu_\x$ and $\tau_\x$) and the denoising rate $\zeta$.
Here we analyze this for few commonly encountered cases. 
A trivial observation from \eqref{eq:convda} is that it is always beneficial to use smaller (or thinner) networks, resulting in faster decay of expected gradients as iterations increase.
This trend, in tandem with observations from \cite{hardt2015train} about generalization imply the superiority of thinner networks -- 
\cite{romero2014fitnets} and others have already shown some empirical evidence for this behavior. As was seen with $1$-NNs, fatter DAs will need large batchsize. 

%%%%%%%%%%%%%%%%%%%%%%%%%%%%%%%%%%%%%%%%%%%%%%%%%%%%%%%%%%%%%%%%%%%%%%%%%%%%%%%%%%%%%%%%%%%%%%%%%%%%%%%%%%%%%%%%%%%%%%%%%%%%%%%%%%%%%%%%%%%%%%%%%%%%%
%%%%%%%%%%%%%%%%%%%%%%%%%%%%%%%%%%%%%%%%%%%%%%%%%%%%%%%%%%%%%%%%%%%%%%%%%%%%%%%%%%%%%%%%%%%%%%%%%%%%%%%%%%%%%%%%%%%%%%%%%%%%%%%%%%%%%%%%%%%%%%%%%%%%%

{\bf Small data moments; ({\it $\mu_\x\approx 0$, $\tau_\x\approx 0$})}  
In this trivial case, there is nothing much to learn. 
$e^{da}$ will be as small as possible, resulting in faster convergence, and the influence of $\zeta$ on \eqref{eq:convda} is nullified as well. % by $\mu_\x ,\tau_\x \approx 0$.
Hence, independent of the complexity of the task, there is no necessity for using large $\zeta$s whenever the input data averages are small. 
The only contributing factor is the network size, and clearly, smaller networks are better. 

%%%%%%%%%%%%%%%%%%%%%%%%%%%%%%%%%%%%%%%%%%%%%%%%%%%%%%%%%%%%%%%%%%%%%%%%%%%%%%%%%%%%%%%%%%%%%%%%%%%%%%%%%%%%%%%%%%%%%%%%%%%%%%%%%%%%%%%%%%%%%%%%%%%%%
%%%%%%%%%%%%%%%%%%%%%%%%%%%%%%%%%%%%%%%%%%%%%%%%%%%%%%%%%%%%%%%%%%%%%%%%%%%%%%%%%%%%%%%%%%%%%%%%%%%%%%%%%%%%%%%%%%%%%%%%%%%%%%%%%%%%%%%%%%%%%%%%%%%%%

{\bf Small denoising rate; ({\it $0 \ll \zeta < 1$})} 
Here $\x * \z \approx \x$, and the noise in gradients $\nabla_{\W}\mathcal{L}(\eta;\W)$ is almost entirely from data statistics. Within this setting, 
\begin{itemize}
\item Small $\mu_\x$ and $\tau_\x$ leads to faster convergence. As they increase, large batches and $N$ are needed. %The trends here are similar to small data moments case discussed earlier.
\item For large $\mu_\x$ and $\tau_\x$, smaller stepsizes and input length ($d_x$) are required to control \eqref{eq:convda}.
In the pathological case where $\mu_\x$, $\tau_\x \sim 1$ and $\zeta \approx 1$, the bounds may be too loose to be relevant.
\end{itemize}
{\it Independent} of how small \eqref{eq:convda} is, large $\zeta$ always leads to overfitting and poor hidden representations \cite{vincent2010stacked}. 
\eqref{eq:convda} predicts this from the convergence perspective. 
To see this, observe that large $\zeta$ (and mid-sized network) implies large expected gradients, and hence, the training time $N$ needs to be reasonably large. 
\cite{hardt2015train} show that networks with large training times may not generalize well. 

%%%%%%%%%%%%%%%%%%%%%%%%%%%%%%%%%%%%%%%%%%%%%%%%%%%%%%%%%%%%%%%%%%%%%%%%%%%%%%%%%%%%%%%%%%%%%%%%%%%%%%%%%%%%%%%%%%%%%%%%%%%%%%%%%%%%%%%%%%%%%%%%%%%%%
%%%%%%%%%%%%%%%%%%%%%%%%%%%%%%%%%%%%%%%%%%%%%%%%%%%%%%%%%%%%%%%%%%%%%%%%%%%%%%%%%%%%%%%%%%%%%%%%%%%%%%%%%%%%%%%%%%%%%%%%%%%%%%%%%%%%%%%%%%%%%%%%%%%%%

{\bf Large denoising rate, ({\it $0 < \zeta \ll 1$})}  
Here $e^{da} \approx \frac{d_xd_h}{16}$. The influence of data statistics is completely nullified by large corruptions 
(i.e., small $\zeta$, see \eqref{eq:loss-da} and the last two terms in \eqref{eq:convdaparam-eda}).
Unless $\mu_\x$, $\tau_\x$ are unreasonably large %(recall that $\mu_\x$, $\tau_\x \leq 1$), 
the convergence is almost entirely controlled by $\zeta$, which in turn would be faster for thinner networks with large batchsize.
%The worst scenario here is $\mu_\x$, $\tau_\x \approx 1$, however, in such a scenario there is nothing much to learn, and the bound becomes moot.

%%%%%%%%%%%%%%%%%%%%%%%%%%%%%%%%%%%%%%%%%%%%%%%%%%%%%%%%%%%%%%%%%%%%%%%%%%%%%%%%%%%%%%%%%%%%%%%%%%%%%%%%%%%%%%%%%%%%%%%%%%%%%%%%%%%%%%%%%%%%%%%%%%%%%
%%%%%%%%%%%%%%%%%%%%%%%%%%%%%%%%%%%%%%%%%%%%%%%%%%%%%%%%%%%%%%%%%%%%%%%%%%%%%%%%%%%%%%%%%%%%%%%%%%%%%%%%%%%%%%%%%%%%%%%%%%%%%%%%%%%%%%%%%%%%%%%%%%%%%
\begin{comment}
{\bf Size of constraint set} 
For small $w_m$, the projected gradients are expected to be small, which is easy to see from \eqref{eq:convdaparam-eda}. 
As $w_m$ increases, large $N$ and/or $B$ are required. 
Since the choice of $w_m$ is independent of other hyper-parameters, all the above described inferences about distributional, structural and learning constants will come into play as $w_m$ increases.
\end{comment}
%%%%%%%%%%%%%%%%%%%%%%%%%%%%%%%%%%%%%%%%%%%%%%%%%%%%%%%%%%%%%%%%%%%%%%%%%%%%%%%%%%%%%%%%%%%%%%%%%%%%%%%%%%%%%%%%%%%%%%%%%%%%%%%%%%%%%%%%%%%%%%%%%%%%%
%%%%%%%%%%%%%%%%%%%%%%%%%%%%%%%%%%%%%%%%%%%%%%%%%%%%%%%%%%%%%%%%%%%%%%%%%%%%%%%%%%%%%%%%%%%%%%%%%%%%%%%%%%%%%%%%%%%%%%%%%%%%%%%%%%%%%%%%%%%%%%%%%%%%%

Beyond these prototypical settings, for small stepsizes $\gamma$, $e^{da}_\gamma$ will be as large as possible making the bound tighter. 
Clearly, the influence of both data moments and denoising rates may be mitigated by increasing $B$ and $N$.
The above described cases are some of the widely used settings, but the interplay from \eqref{eq:convda} and \eqref{eq:convdaparam-eda} in Corollary \ref{thm:expgradda} is much more involved. 
The trends and interpretations derived here will be useful in understanding multi-layer networks. 
Recall the discussion about large deviation estimates from Corollary \ref{thm:conv1nn}. 
A similar such result relating $N$ to the number of instances $S$ and $B$ can be obtained for DA as well, however, 
due to space restrictions we omit it here. %defer it to the longer version of the paper.

\section{Multi-layer Networks}
\label{sec:multi}

We now extend our analysis to multi-layer networks. 
Using Theorem \ref{thm:expgradda} as a starting point, we first consider a $L$-NN that performs {\it layer-wise pretraining} using DAs from Section \ref{sec:pretrain} 
before backpropagation based supervised finetuning.
Since the layer-wise pretraining uses box-constraints, we are still interested in the expected projected gradients, accumulated across all the layers.
The resulting bound allows us to incorporate feature dropout (during the supervised tuning stage) in both the input and hidden layers later in Section \ref{sec:multidrop}.  
Recall the discussion about $L$-NN learning mechanisms and the corresponding loss functions (see \eqref{eq:loss-mulnn}) from Section \ref{sec:prelim}.
A consequence of this general result is a clear and intuitive relation between dropout based supervised learning, layer-wise pretraining and other structural and distributional parameters.
Alg. 3 in the technical report presents this multi-layer RSG procedure.
For notational convenience we collectively denote the final estimate $\W^{R}_{1},\ldots,\W^{R}_{L}$ simply as $\W^R$ in the results. 

%%%%%%%%%%%%%%%%%%%%%%%%%%%%%%%%%%%%%%%%%%%%%%%%%%%%%%%%%%%%%%%%%%%%%%%%%%%%%%%%%%%%%%%%%%%%%%%%%%%%%%%%%%%%%%%%%%%%%%%%%%%%%%%%%%%%%%%%%%%%%%%%%%%%%%%%%%%%
%%%%%%%%%%%%%%%%%%%%%%%%%%%%%%%%%%%%%%%%%%%%%%%%%%%%%%%%%%%%%%%%%%%%%%%%%%%%%%%%%%%%%%%%%%%%%%%%%%%%%%%%%%%%%%%%%%%%%%%%%%%%%%%%%%%%%%%%%%%%%%%%%%%%%%%%%%%%

\subsection{Layer-wise Pretraining}
\label{sec:multipretrain}

The following result shows the decay of expected projected gradients for multi-layer RSG.  
$D_f$ here denotes the initial deviation of the objective {\it after} the $L-1$ layers have been pretrained.
$e^m_l$ for $l=1,\ldots,L$ encode the structural and learning hyper-parameters of the network, 
and we assume that all the hidden layers are pretrained to the same degree i.e., each of the $L-1$ layers are pretrained to a given $(\alpha,\delta_{\alpha})$ solution.
\begin{corollary}[{\bf Multi-layer Network}] \label{thm:expgradmulnn}
Consider a multi-layer RSG with {\it no} dropout learned via layer-wise box-constrained DA pretraining followed by supervised backpropagation with constant stepsizes $\gamma_l\forall l$.
Let $e^m_1 = \frac{\gamma_1}{4}d_0d_1d_2w^l_m$, $e^m_l = \frac{\gamma_l}{4}d_{l-1}d_{l}d_{l+1}w^l_m$ and $e^m_L = \frac{13d_{L-1}d_L\gamma^2_L}{256}$.
Whenever $\gamma_l < \frac{20}{\alpha d_{l+1}w^l_m}$, and the hidden layers are pretrained for a $(\alpha,\delta_{\alpha})$ solution (as defined in Definition \ref{thm:epsdel}), we have 
\begin{equation} \label{eq:convmulnn}
\mathbb{E} \| \nabla_{\W} \tf(\W^R) \|^2  \leq \frac{1}{e^m_\gamma} \left( \frac{D_f}{N} + \frac{1}{B} (e^m_L + \alpha\sum_{l=1}^{L-1} e^m_l) \right)
\end{equation}
where $e^m_\gamma = \min \left\{ \gamma_L - \frac{13}{16}(\gamma_L)^2, \gamma_l - \frac{\alpha d_{l+1}w^l_m}{20}(\gamma_l)^2 \right\}$
\end{corollary}

%%%%%%%%%%%%%%%%%%%%%%%%%%%%%%%%%%%%%%%%%%%%%%%%%%%%%%%%%%%%%%%%%%%%%%%%%%%%%%%%%%%%%%%%%%%%%%%%%%%%%%%%%%%%%%%%%%%%%%%%%%%%%%%%%%%%%%%%%%%%%%%%%%%%%%%%%%%%
%%%%%%%%%%%%%%%%%%%%%%%%%%%%%%%%%%%%%%%%%%%%%%%%%%%%%%%%%%%%%%%%%%%%%%%%%%%%%%%%%%%%%%%%%%%%%%%%%%%%%%%%%%%%%%%%%%%%%%%%%%%%%%%%%%%%%%%%%%%%%%%%%%%%%%%%%%%%

{\it Remarks:} 
The structure of \eqref{eq:convmulnn} is similar to those from \eqref{eq:conv1nn} and \eqref{eq:convda}.
Hence, the trends suggested by the interplay of $D_f$, $L$ and $d_0,\ldots,d_L$, stepsizes, $N$ and $B$ are similar to those observed from the Theorems \ref{thm:expgrad} and \ref{thm:expgradda}. 
However, as expected, the interactions are much more complex because of the presence of multiple layers.
For fixed network lengths and stepsizes, $e^m_l$ are constants, and encode the variance of gradients within the $l^{th}$ layer and the corresponding free parameters.
As network size increases, these constants increase proportionally, there by requiring large $N$ and $B$. 
The observations about thinner networks being superior to fatter ones can also be seen from $e^m_l$s, 
and \eqref{eq:convmulnn} suggests that the network depth should {\it not} be more than necessary. 
Such a minimum depth would depend on the trade-off of convergence (from \eqref{eq:convmulnn}) and generalization \cite{hardt2015train, bengio2009learning}.

%%%%%%%%%%%%%%%%%%%%%%%%%%%%%%%%%%%%%%%%%%%%%%%%%%%%%%%%%%%%%%%%%%%%%%%%%%%%%%%%%%%%%%%%%%%%%%%%%%%%%%%%%%%%%%%%%%%%%%%%%%%%%%%%%%%%%%%%%%%%%%%%%%%%%%%%%%%%
%%%%%%%%%%%%%%%%%%%%%%%%%%%%%%%%%%%%%%%%%%%%%%%%%%%%%%%%%%%%%%%%%%%%%%%%%%%%%%%%%%%%%%%%%%%%%%%%%%%%%%%%%%%%%%%%%%%%%%%%%%%%%%%%%%%%%%%%%%%%%%%%%%%%%%%%%%%%

The influence of layer-wise DAs are concealed within $\alpha$ and $\delta_{\alpha}$ and the corresponding $D_f$.
A small $\alpha$ in \eqref{eq:convmulnn} (which controls the goodness of pretraining) suggests that the influence of the $\h^1,\ldots,\h^{L-1}$ on backpropagation tuning is very small. 
In such a scenario, the tuning is mostly confined to mapping $\h^{L-1}$ to the outputs $\y$ -- 
which is not necessarily a good thing, and one would want to allow for all $\h^l$s (and $\W_l$s) to change if needed \cite{vincent2010stacked, saxe2011random}. 
On the other hand, a large $D_f$ controls the alternate regime where $\h^l$s are not ``good enough'' to abstractly represent the input data.
In such a case, aggressive tuning is needed, as suggested by \eqref{eq:convmulnn}, with large $N$ and $B$.
Overall, \eqref{eq:convmulnn} implies that the goodness (or badness) of pretraining will be passed on to the tuning stage convergence via $\alpha$ and $D_f$.
As suggested earlier, these trade-offs can be related to the training times (proportional to $N$, derived from generalization \cite{hardt2015train}).
Recently, \cite{yosinski2014transferable} showed empirical evidence that aggressive pretraining (especially in higher layers) 
results in $\h^l$s that may not be transferable across multiple arbitrary tasks. 
Theorem \ref{thm:expgradmulnn} suggests the same -- $\alpha$ (and $D_f$) should not be reasonably small to allow for aggressive tuning to arbitrary learning tasks. %(i.e., different types of $\y$s).
%
%%%%%%%%%%%%%%%%%%%%%%%%%%%%%%%%%%%%%%%%%%%%%%%%%%%%%%%%%%%%%%%%%%%%%%%%%%%%%%%%%%%%%%%%%%%%%%%%%%%%%%%%%%%%%%%%%%%%%%%%%%%%%%%%%%%%%%%%%%%%%%%%%%%%%%%%%%%%
%%%%%%%%%%%%%%%%%%%%%%%%%%%%%%%%%%%%%%%%%%%%%%%%%%%%%%%%%%%%%%%%%%%%%%%%%%%%%%%%%%%%%%%%%%%%%%%%%%%%%%%%%%%%%%%%%%%%%%%%%%%%%%%%%%%%%%%%%%%%%%%%%%%%%%%%%%%%
%
It is clear from \eqref{eq:convmulnn} that we can control the {\it convergence} of multi-layer nets by preceding the backpropagation tuning with layer-wise pretraining.
There is already very strong empirical evidence for the generalization performance of pretraining \cite{erhan2010does, vincent2010stacked, saxe2011random}. %, bengio2007greedy}. 
Corollary \ref{thm:expgradmulnn} complements these studies with guaranteed convergence of gradients. 

\section{Multi-layer with Dropout}
\label{sec:multidrop}

Using ReLUs and dropout, \cite{nair2010rectified, krizhevsky2012imagenet, szegedy2014going} and others have shown that whenever large amounts of labeled data is available, 
pretraining might not be {\em necessary} to achieve good generalization, clearly suggesting an underlying relationship between dropout and layer-wise pretraining.
The following result summarizes the convergence of expected projected gradients in this general setting where the $L-1$ layers are not pretrained, 
and instead dropout is induced in input and all hidden layers, 
and supervised backproporagion is performed directly with random initializations for $\W^{1}_{1},\ldots,\W^{1}_{L}$ \cite{srivastava2014dropout}.
Here, $\zeta$ denotes the dropout rate for all layers i.e., 
w.p. $1-\zeta$ a unit is dropped and all parameters corresponding to this unit are not updated (see Section \ref{sec:prelim}, and \eqref{eq:loss-mulnn}).
It is reasonable to expect some interaction between $\alpha$ (the pretraining goodness, see Corollary \ref{thm:expgradmulnn}) and $\zeta$. 

%%%%%%%%%%%%%%%%%%%%%%%%%%%%%%%%%%%%%%%%%%%%%%%%%%%%%%%%%%%%%%%%%%%%%%%%%%%%%%%%%%%%%%%%%%%%%%%%%%%%%%%%%%%%%%%%%%%%%%%%%%%%%%%%%%%%%%%%%%%%%%%%%%%%%%%%%%%%
%%%%%%%%%%%%%%%%%%%%%%%%%%%%%%%%%%%%%%%%%%%%%%%%%%%%%%%%%%%%%%%%%%%%%%%%%%%%%%%%%%%%%%%%%%%%%%%%%%%%%%%%%%%%%%%%%%%%%%%%%%%%%%%%%%%%%%%%%%%%%%%%%%%%%%%%%%%%

\begin{corollary}[{\bf Pretraining vs. Dropout}] \label{thm:pretvsdrop}
Given the input and hidden layer dropout rate $\zeta$, for learning the $L$-layered network from Corollary \ref{thm:expgradmulnn} 
that is pretrained to a $(\alpha,\delta_{\alpha})$ solution, we have
\begin{equation} \begin{aligned} \label{eq:mlnnpretdrop}
\mathbb{E}_{R,\eta} &\| \nabla_{\W} \tf(\W^R) \|^2 \leq \frac{D_f}{Ne^m_\gamma\zeta^2} \\
&+ \frac{1}{e^m_\gamma B} \left( \frac{e^m_L}{\zeta} + \alpha e^m_{L-1} + \alpha \zeta \sum_{l=1}^{L-2} e^m_l \right)
\end{aligned} \end{equation}
\end{corollary}
{\it Remarks:} This is our most general result.
Although, the schools of layer-wise pretraining and dropout have been studied independently, from our knowledge, 
this is the first theoretical result that explicitly combines these two regimes in a systematic way.
Recall that $e^m_1,\ldots,e^m_L$ encode $L$-NN's degrees of freedom.
Hence, the first term in \eqref{eq:mlnnpretdrop} corresponds to the outputs' goodness-of-fit, 
while the other terms represent the effective `reduction' in the number of free parameters because of pretraining.
Independent of $\zeta$, \eqref{eq:mlnnpretdrop} clearly implies that pretraining always improves convergence (since $\alpha$ will reduce). 
\cite{lee2009convolutional, srivastava2014dropout} have shown empirically that this is especially true for improving generalization.
To better interpret \eqref{eq:mlnnpretdrop}, consider the two standard mechanisms --- dropout learning with and without layer-wise pretraining. 
Given $L$, $d_0,\ldots,d_L$ and stepsizes ($\gamma_1,\ldots,\gamma_L$), $e^m_1,\ldots,e^m_L$ from Corollary \ref{thm:expgradmulnn} are constants. 
We assume $D_f$ (which depends on pretraining) to be reasonably large because the loss is calculated over predicted vs. observed $\y$s (see \eqref{eq:loss-mulnn}).

%%%%%%%%%%%%%%%%%%%%%%%%%%%%%%%%%%%%%%%%%%%%%%%%%%%%%%%%%%%%%%%%%%%%%%%%%%%%%%%%%%%%%%%%%%%%%%%%%%%%%%%%%%%%%%%%%%%%%%%%%%%%%%%%%%%%%%%%%%%%%%%%%%%%%%%%%%%%
%%%%%%%%%%%%%%%%%%%%%%%%%%%%%%%%%%%%%%%%%%%%%%%%%%%%%%%%%%%%%%%%%%%%%%%%%%%%%%%%%%%%%%%%%%%%%%%%%%%%%%%%%%%%%%%%%%%%%%%%%%%%%%%%%%%%%%%%%%%%%%%%%%%%%%%%%%%%

\subsection{ Pretraining + Dropout}\label{sec:pret+drop}

\begin{itemize}
\item If $L$-NN is pretrained to a small $\alpha$, the last two terms in \eqref{eq:mlnnpretdrop} are already as small as they can be, and the rest will dominate them.
For a given $B$ and $N$, the best choice for $\zeta$ will then be $\approx 1$ i.e., very minimal or {\it no} dropout. 
This is essentially ``good'' layer-wise pretraining followed by supervised fine-tuning, which is known to work well \cite{hinton2006reducing, erhan2010does}.
Hence, \eqref{eq:mlnnpretdrop} presented a succinct and  clear theoretical argument for the classical revival of deep networks from the perspective of convergence (and eventually generalization).
\item Alternatively, if the pretraining is poor (i.e., large $\alpha$), and we still operate in the $\zeta \to 1$ regime, the fine-tuning will update the full network in each iteration.
This would result in overfitting -- the fundamental argument that necessitates dropout (empirically made clear in \cite{srivastava2014dropout, wager2013dropout}).
Hence one needs to decrease $\zeta$, resulting in slower convergence because the first two terms will rapidly increase as $\zeta$ decreases. 
This is expected since dropout essentially adds `noise' to the solution path, 
forcing backpropagation to work with a subset of all activations \cite{baldi2014dropout, wager2013dropout}, and overall, involving more work.
\end{itemize}

%%%%%%%%%%%%%%%%%%%%%%%%%%%%%%%%%%%%%%%%%%%%%%%%%%%%%%%%%%%%%%%%%%%%%%%%%%%%%%%%%%%%%%%%%%%%%%%%%%%%%%%%%%%%%%%%%%%%%%%%%%%%%%%%%%%%%%%%%%%%%%%%%%%%%%%%%%%%
%%%%%%%%%%%%%%%%%%%%%%%%%%%%%%%%%%%%%%%%%%%%%%%%%%%%%%%%%%%%%%%%%%%%%%%%%%%%%%%%%%%%%%%%%%%%%%%%%%%%%%%%%%%%%%%%%%%%%%%%%%%%%%%%%%%%%%%%%%%%%%%%%%%%%%%%%%%%

\subsection{Only Dropout}\label{sec:drop-only}

With no pretraining, there is a complex trade-off between the terms involving $\zeta$ in \eqref{eq:mlnnpretdrop}.
The optimal $\zeta$ will need to balance out the variance from the hidden layers (the last term) and the goodness of output approximation (first and second terms).
For certain values of $D_f$ and $e^m_l$'s, the best $\zeta$ will be $\sim 0.5$, which was recently reported as the best rate as per dropout dynamics \cite{baldi2014dropout}.
Large values of $B$ and $N$ will ensure that the bound is small. 
Large $N$, in turn, implies larger training dataset size (see the setup from Corollary \ref{thm:conv1nn} and \ref{thm:expgradmulnn}).
Putting it another way, if large amounts of labeled data is available, one can by-pass pretraining completely and only perform supervised backpropagation 
forcing all the terms in \eqref{eq:mlnnpretdrop} to reduce with reasonably large $B$ and number of epochs.
This was the argument put forth by the recent school of deep learning \cite{krizhevsky2012imagenet, szegedy2014going} -- %ciregan2012multi
fully supervised dropout with very large amounts of training data -- where pretraining has been for the most part neglected completely.

%%%%%%%%%%%%%%%%%%%%%%%%%%%%%%%%%%%%%%%%%%%%%%%%%%%%%%%%%%%%%%%%%%%%%%%%%%%%%%%%%%%%%%%%%%%%%%%%%%%%%%%%%%%%%%%%%%%%%%%%%%%%%%%%%%%%%%%%%%%%%%%%%%%%%%%%%%%%
%%%%%%%%%%%%%%%%%%%%%%%%%%%%%%%%%%%%%%%%%%%%%%%%%%%%%%%%%%%%%%%%%%%%%%%%%%%%%%%%%%%%%%%%%%%%%%%%%%%%%%%%%%%%%%%%%%%%%%%%%%%%%%%%%%%%%%%%%%%%%%%%%%%%%%%%%%%%

Overall, Corollary \ref{thm:pretvsdrop} corroborates many existing observations about pretraining and dropout, and provides new tools to {\em guide} the learning procedure, in general.
It guarantees convergence, and allows us to {\it explicitly} calculate the best possible settings for all hyper-parameters, 
to achieve a certain level of generalization or training time \cite{hardt2015train}.
A more extensive discussion about the outcomes and insights from the interplay in Corollary \ref{thm:pretvsdrop} will be included in the longer version of the paper.
The three algorithms presented in this work are only minor modifications over the classical autoencoder and backpropagation pipeline, and so, are straightforward to implement.

%%%%%%%%%%%%%%%%%%%%%%%%%%%%%%%%%%%%%%%%%%%%%%%%%%%%%%%%%%%%%%%%%%%%%%%%%%%%%%%%%%%%%%%%%%%%%%%%%%%%%%%%%%%%%%%%%%%%%%%%%%%%%%%%%%%%%%%%%%%%%%%%%%%%%%%%%%%%
%%%%%%%%%%%%%%%%%%%%%%%%%%%%%%%%%%%%%%%%%%%%%%%%%%%%%%%%%%%%%%%%%%%%%%%%%%%%%%%%%%%%%%%%%%%%%%%%%%%%%%%%%%%%%%%%%%%%%%%%%%%%%%%%%%%%%%%%%%%%%%%%%%%%%%%%%%%%

\section{Experiments}
\label{sec:exps}
 
\begin{figure*}[!ht]\centering
\subfloat[]{\includegraphics[width=33mm]{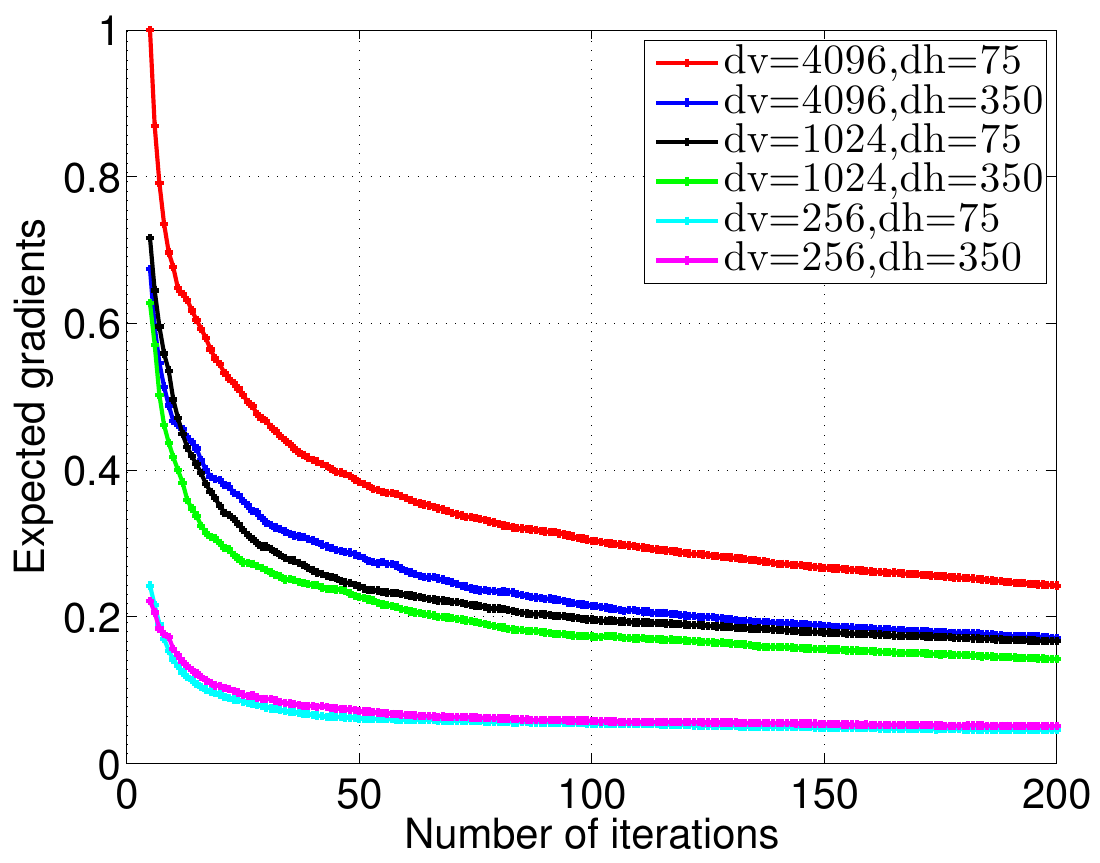}} 
\subfloat[]{\includegraphics[width=33mm]{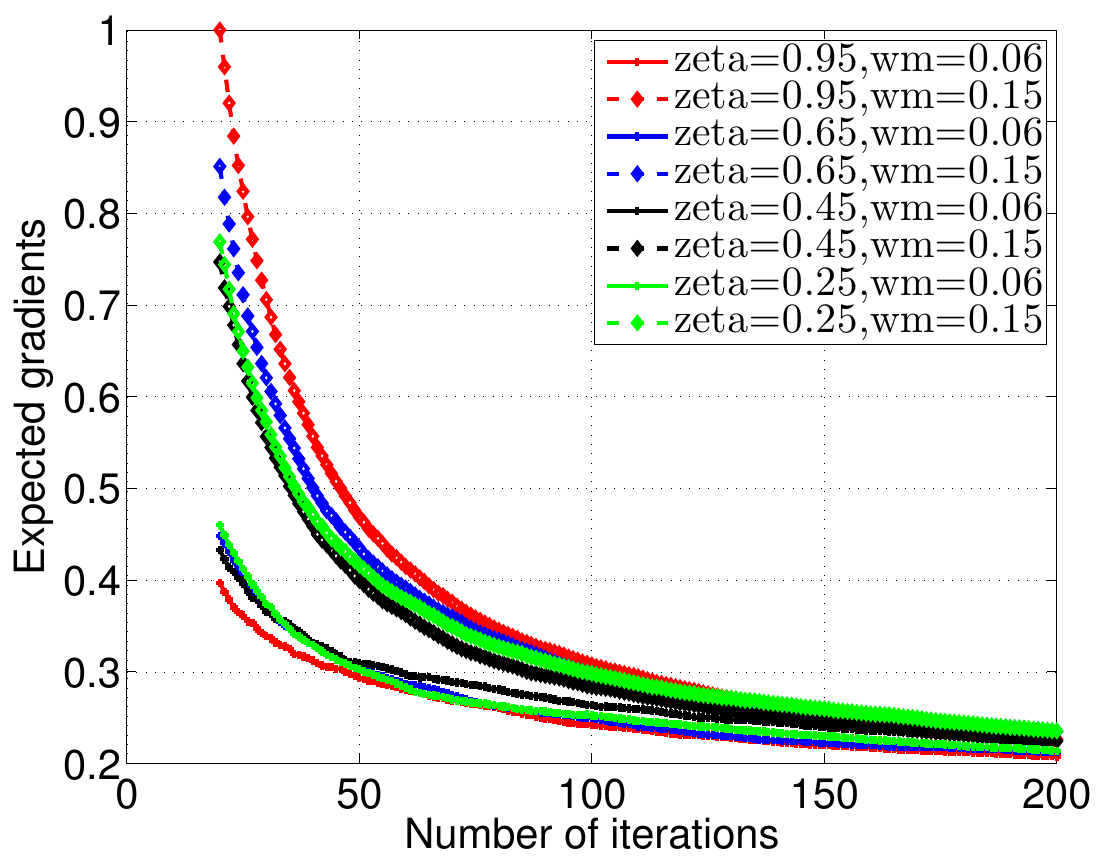}}
\subfloat[]{\includegraphics[width=33mm]{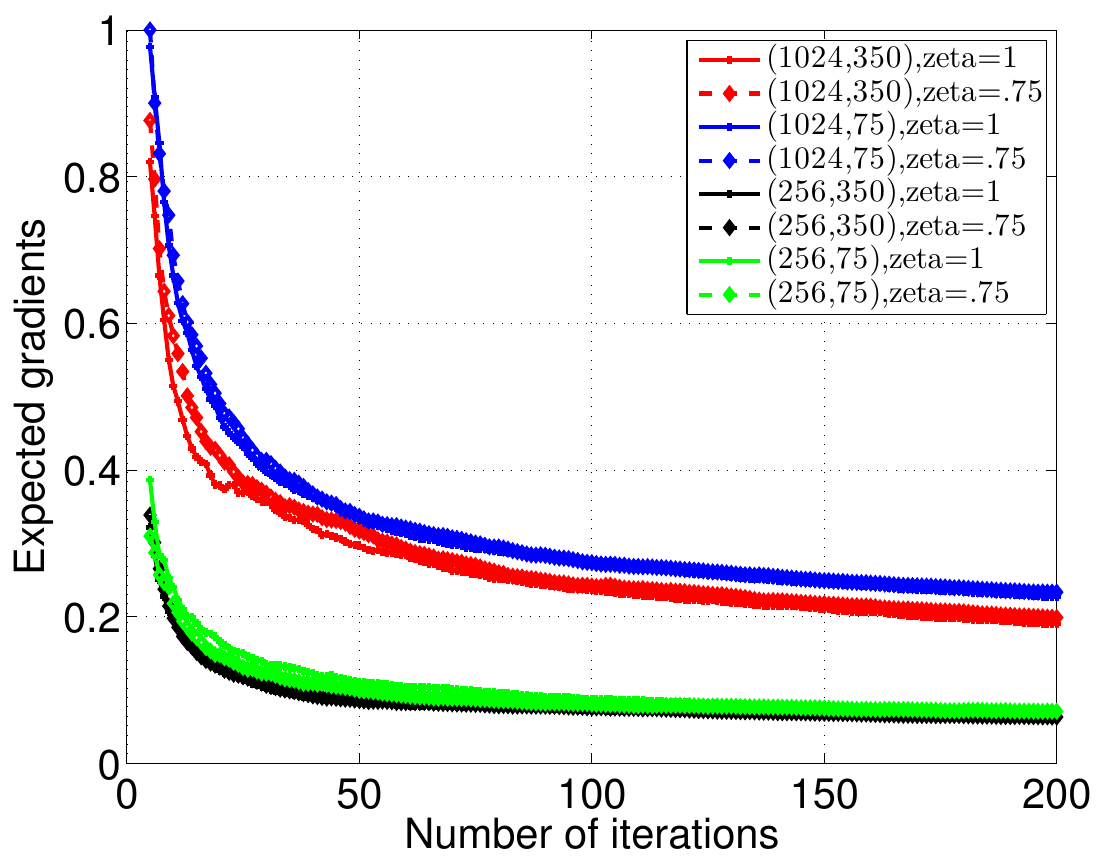}} 
\subfloat[]{\includegraphics[width=33mm]{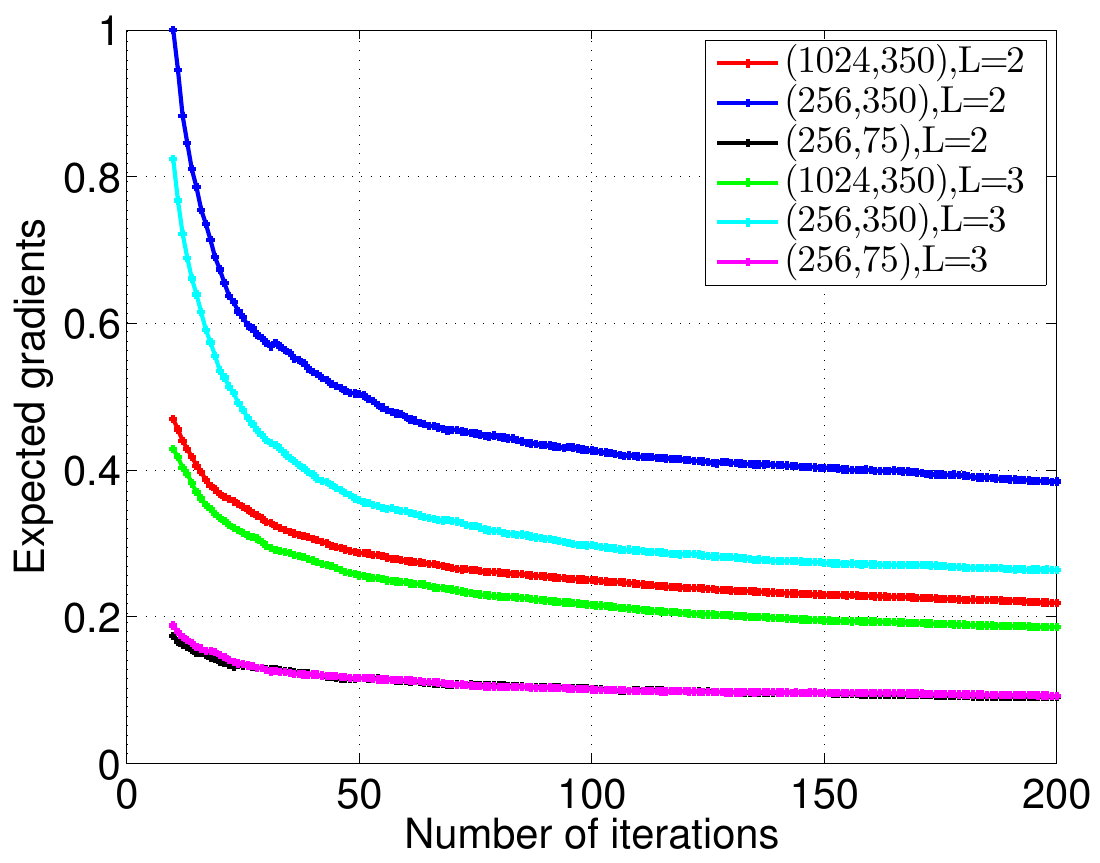}} 
\subfloat[]{\includegraphics[width=33mm]{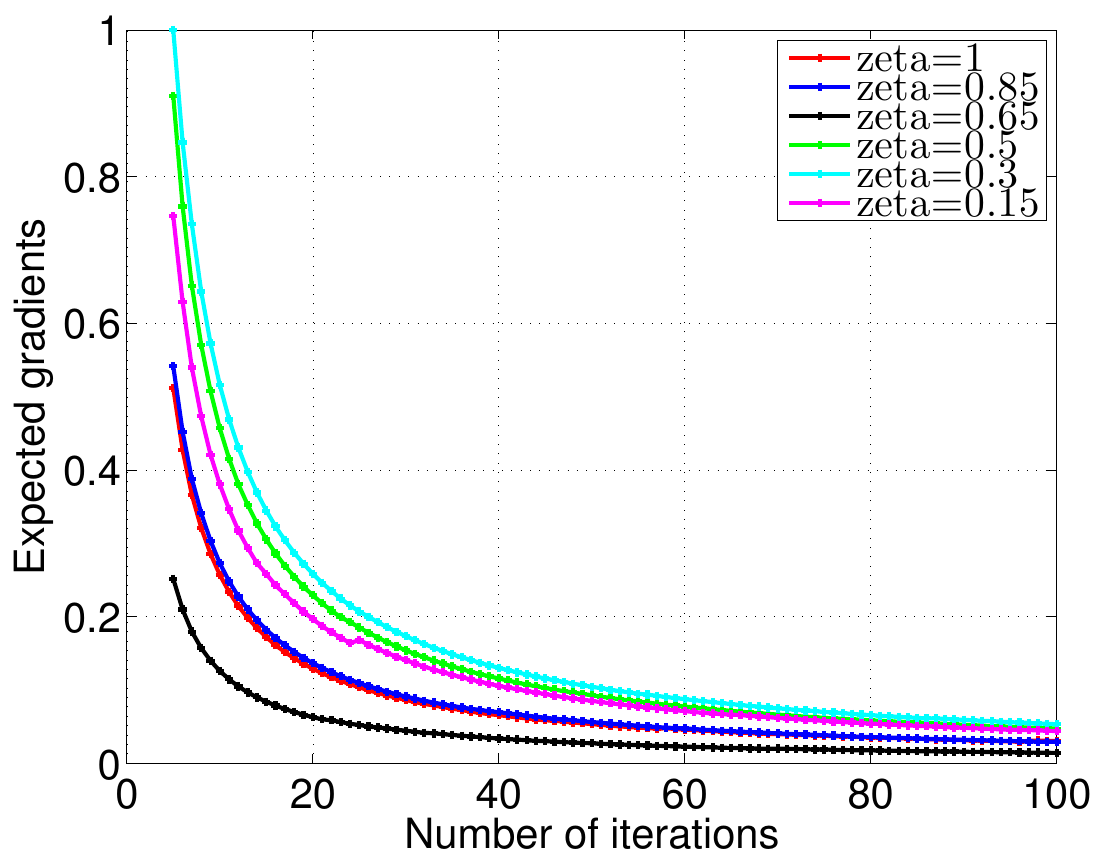}} \vspace{-5pt}
\caption{\footnotesize \label{fig:allexp} Expected gradients (CIFAR) for pretraining (a,b) and multi-layer net (c,d). (e) Influence of dropout.
$y$-axes are scaled to maximum. $B=100$.}
\end{figure*}

We evaluate the bounds in \eqref{eq:convda}, \eqref{eq:convmulnn} and \eqref{eq:mlnnpretdrop} using MNIST, CIFAR-10, CALTECH-101 and a brain imaging dataset.
Figures \ref{fig:allexp} summarizes a fraction of the findings on CIFAR-10. The longer version of the paper will have complete set of evaluations.
%{\bf Convergence vs. $d_x$, $d_h$, $w_m$, $\zeta$: }
Figure \ref{fig:allexp}(a) shows the expected gradients vs. network sizes. As predicted in \eqref{eq:conv1nn}, the convergence is slower for large networks. 
Visible and hidden layer lengths have unequal influence on the convergence (see remarks of Corollary \ref{thm:expgradda}).
The influence of denoising rate and $w_m$ is in Figure \ref{fig:allexp}(b).
The $y$-axis shows the expectation of {\it projected} gradients on $[-w_m,w_m]$, and as suggested by \eqref{eq:convda}, the convergence is faster for small $w_m$s.
It is interesting that the choice of $\zeta$s has an almost negligible influence (refer Figure \ref{fig:allexp}(b), for a given $w_m$. 
Figure \ref{fig:allexp}(c), shows the interaction of network lengths vs $\zeta$, 
and as observed from Figure \ref{fig:allexp}(a), the networks lengths dominate the convergence with visible layer length being the most influential factor. %(also see Figure \ref{fig:allexp}(a,c)).
The plots in Figure \ref{fig:allexp} correspond to the data moments $\mu_\x\sim 0.5$ and $\tau_\x\sim 0.3$, 
implying that the terms in $e^{da}$ (refer \eqref{eq:convdaparam-eda}) are non-negligible.
Figure \ref{fig:allexp}(d) shows the influence of the network depth.
Clearly, the expected gradients are influenced more by layer lengths than the number of layers itself \eqref{eq:convmulnn}.
Figure \ref{fig:allexp}(e) shows the effect of changing the dropout rate in a $3$-layered network. 
Although the overall effect is small, the convergence is slower for very small (red curve) and very large (cyan, magenta curves) $\zeta$s (see \eqref{eq:mlnnpretdrop}).

%%%%%%%%%%%%%%%%%%%%%%%%%%%%%%%%%%%%%%%%%%%%%%%%%%%%%%%%%%%%%%%%%%%%%%%%%%%%%%%%%%%%%%%%%%%%%%%%%%%%%%%%%%%%%%%%%%%%%%%%%%%%%%%%%%%%%%%%%%%%%%%%%%%%%%%%%%%%
%%%%%%%%%%%%%%%%%%%%%%%%%%%%%%%%%%%%%%%%%%%%%%%%%%%%%%%%%%%%%%%%%%%%%%%%%%%%%%%%%%%%%%%%%%%%%%%%%%%%%%%%%%%%%%%%%%%%%%%%%%%%%%%%%%%%%%%%%%%%%%%%%%%%%%%%%%%%

{\footnotesize \vspace{2mm}\noindent {\bf Acknowledgements:} We thank the anonymous reviewers. 
We are supported by NIH AG040396, NSF CAREER 1252725, NSF CCF 1320755, the UW grants ADRC AG033514, ICTR 1UL1RR025011 and CPCP AI117924.}

{\scriptsize
\bibliographystyle{IEEEtran}
\bibliography{dlconv-allerton-cam}}

\end{document}